\documentclass[acmsmall,screen,nonacm]{acmart}
\setcopyright{none}
\renewcommand\footnotetextcopyrightpermission[1]{}

\AtBeginDocument{}

\usepackage{enumitem}
\usepackage{multirow}
\usepackage{tikz}
\usetikzlibrary{arrows.meta}
\usepackage{tabularx}
\AtEndEnvironment{algocf}{\vspace{-8pt}}
\usepackage[linesnumbered,ruled,vlined]{algorithm2e}
\usepackage{amsthm}
\theoremstyle{definition}
\newtheorem{definition}{Definition}

\begin{document}

\title[Hierarchical Feedback-Guided LLM Planning for CPSs]
{From Natural Language Intent to Executable Control: Hierarchical Feedback-Guided LLM Planning for Multi-component Cyber-Physical Systems}

\author{Jinglin Dai}
\email{daijinglin@smail.nju.edu.cn}
\affiliation{\institution{Nanjing University}
  \city{Nanjing}
  \country{China}
}

\author{Jiawan Wang}
\email{wjwnju@nju.edu.cn}
\affiliation{\institution{Nanjing University}
  \city{Nanjing}
  \country{China}
}

\author{Ruiqi Hong}
\email{rqhong@smail.nju.edu.cn}
\affiliation{\institution{Nanjing University}
  \city{Nanjing}
  \country{China}
}

\author{Lei Bu}
\email{bulei@nju.edu.cn}
\affiliation{\institution{Nanjing University}
  \city{Nanjing}
  \country{China}
}

\begin{abstract}
Large language models (LLMs) provide a convenient interface for translating natural language commands into task plans for cyber-physical systems (CPSs).
However, two gaps hinder their use for real execution: generated plans may misinterpret or omit the user’s intent, and even a logically plausible plan may be infeasible under component dynamics, interactions, or safety constraints.
Existing LLM-based planning approaches often lack informative feedback on why a generated plan fails, providing little guidance for revising the plan accordingly.

To address these challenges, we present a hierarchical feedback framework that uses different types of feedback information to guide LLMs in generating executable plans for CPSs.
We prompt LLMs to generate both a structured task plan and corresponding signal temporal logic (STL) constraints, and design a sound algorithm that identifies definite contradictions between them and provides feedback for revising inconsistencies.
At the execution level, when control synthesis fails, we extract different types of diagnostic information, such as violated constraints and conflicting component assignments, and feed them back to guide task replanning.
Experiments on seven heterogeneous multi-component CPS scenarios have shown that our framework can generate executable plans and feasible control executions that satisfy the original natural language missions with a success rate of 87.9\% with GPT-5.

\end{abstract}



\maketitle

\newcommand{\seq}{\mathbin{;}}
\section{Introduction}
\label{sec:intro}
Cyber-physical systems (CPSs), which consist of multiple components with different capabilities, are increasingly expected to carry out complex missions in open physical environments.
Preparing task plans and inputs for control synthesis for these missions requires knowledge of the component models and the ability to express requirements formally.
This creates a need for interfaces that allow users without such expertise to describe missions in natural language and translate those descriptions into plans from which the underlying tools can synthesize feasible executions.

Recent work~\cite{li_large_2026, kannan_smartllm_2024, xu_nl2hltl2plan_2025, wu_selp_2025, chen_autotamp_2024} has explored large language models (LLMs) as natural language interfaces for multi-agent task allocation and planning.
Given a natural language command, LLMs can be prompted to decompose the mission into subtasks and assign them to different agents for downstream execution.
Some work~\cite{kannan_smartllm_2024} then turns the resulting plan into executable code that runs the subtasks sequentially or in parallel in a simulator.
Other work ~\cite{chen_autotamp_2024} uses LLMs to translate intent into intermediate representations, such as signal temporal logic (STL) formulas, which specify task goals and constraints for downstream planners.
However, some approaches~\cite{kannan_smartllm_2024, xu_nl2hltl2plan_2025} do not include a repair loop after failure, while others~\cite{chen_autotamp_2024} primarily use feedback to correct syntactic and semantic errors in STL translation.
Effective repair must address both errors in the generated task structure and failures revealed when that structure is used for the underlying systems.
Two challenges remain in generating executable plans for CPSs:

\noindent \textbf{Challenge 1:} A semantic gap exists between a mission expressed in natural language and the task plan LLMs generate from it.
Requirements on task ordering and coordination may be omitted or misinterpreted during task decomposition.
A key challenge, therefore, is to expose errors in the translation of mission requirements into a task structure before attempting control synthesis.

\noindent \textbf{Challenge 2:} A feasibility gap may arise between a task plan and a feasible execution in the target CPS.
Even when the plan captures the mission requirements, its execution depends on component dynamics, interactions among components, and physical and temporal constraints.
Some conflicts become apparent only during simulations or control synthesis.
Recovering from such failures may require changes to task decomposition or component allocation.
However, knowing that synthesis failed does not indicate which planning decisions should be reconsidered.

For challenge 1, we prompt LLMs to produce two complementary representations of the same mission: a structured task plan that organizes how subtasks are carried out, and task-level STL specifications that express temporal requirements on their execution.
This makes it possible to expose conflicts between the proposed task structure and the requirements expressed in the specifications.
We design a novel consistency checking algorithm to detect contradictions between the task plan and the STL specifications.
Detected contradictions provide feedback for revising the plan or specifications before control synthesis.

For challenge 2, we extract diagnostic information from the underlying tool to guide replanning.
When synthesis fails, we analyze the failure logs to identify the probable cause.
We distinguish among several types of failures and extract corresponding diagnostic information, such as violations of physical constraints and conflicts caused by assigning the same component to parallel subtasks.
We return this information to the task plan generation stage so that the LLM can revise the task decomposition and component allocation while taking the system’s physical behavior into account.

The main contributions of this work are as follows:
\begin{itemize}[leftmargin=1.5em]
\item[--] We design an algorithm to check the consistency between the task plans and STL specifications generated by LLMs.
It detects contradictions without constructing an execution, enabling repair before control synthesis.
(See Sect.~\ref{sec:taskallocation})
\item[--] We extract control synthesis diagnostics to guide task replanning.
It links control synthesis failures to violated constraints, component behaviors, and affected tasks, providing diagnostic information for revising task decomposition and allocation.
(See Sect.~\ref{sec:controlgeneration})
\item[--] We implement the framework and evaluate it across seven heterogeneous multi-component CPS scenes.
The framework achieves an overall success rate of 87.9\% with GPT-5.
(See Sect.~\ref{sec:exp})
\end{itemize}

\section{Background}

\subsection{Preliminaries}
\label{sec:preliminaries}
This section introduces the concepts used throughout the paper.
Section~\ref{sec:motivating-example} presents a detailed example that illustrates each of them.
We begin with hybrid automata (HAs), which we use to model each component of a CPS.
HAs capture both discrete mode transitions and continuous dynamics.

\begin{figure}[!t]
  \centering
  \vspace{-0.3cm}
  \includegraphics[width=0.75\textwidth]{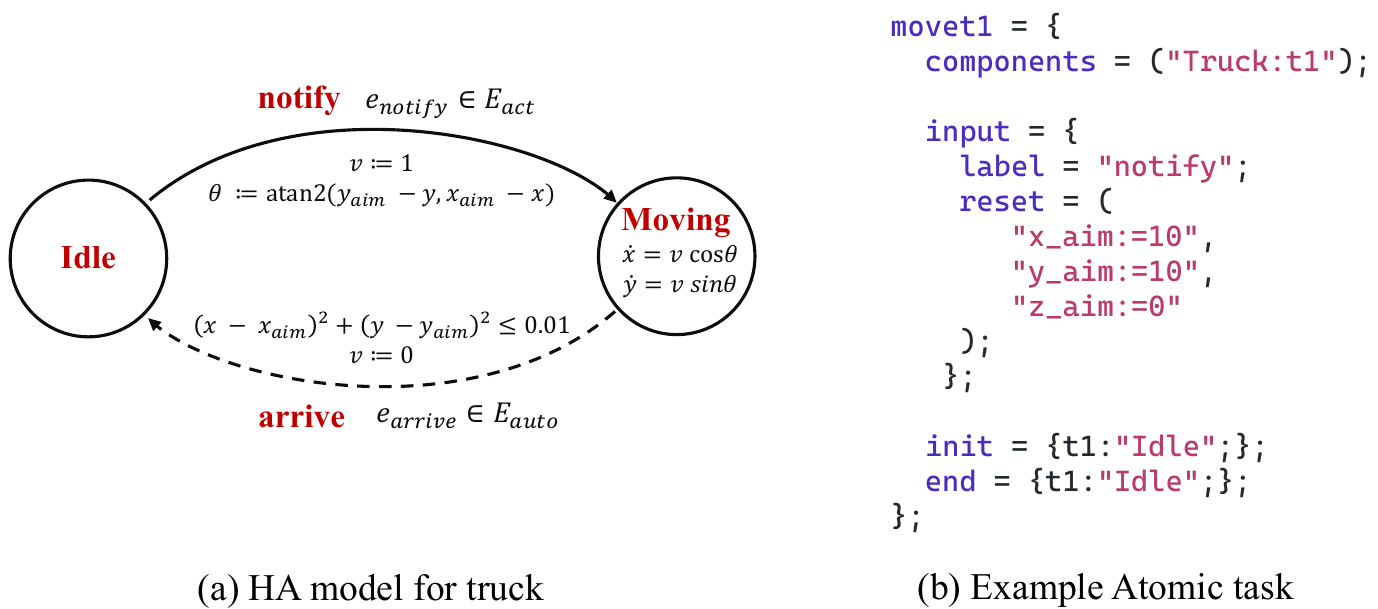}
  \vspace{-0.3cm}
  \caption{Hybrid automaton model of a truck component and an example atomic task. }
  \Description{Truck hybrid automaton and an example atomic task configuration. }
  \label{HA}
  \vspace{-0.5cm}
\end{figure}

\begin{definition}
\textbf{(Hybrid Automaton~\cite{Henzinger_HA_1996})}.
A hybrid automaton is a tuple
\[
H=(X,Q,F,Inv,E,G,R),
\]
where $X$ is a set of real-valued variables, $Q$ is a set of discrete modes, $F_q$ defines the continuous flow in mode $q$, $Inv_q$ is the invariant of mode $q$, $E\subseteq Q\times Q$ is a set of discrete transitions, $G_e$ and $R_e$ are the guard and reset of transition $e$, respectively.

The state of $H$ is a pair $(q,x)\in Q\times\mathbb{R}^{|X|}$.
It can evolve continuously inside the current mode $q$ according to $F_q$ as long as $Inv_q$ holds, or jump discretely through a transition $e$ when its guard $G_e$ is satisfied, after which variables are updated by the reset $R_e$.

We further partition the transition set $E$ into activated transitions $E_{act}$ and automatic transitions $ E _ {auto}$, with $E = E_{act} \cup E _ {auto}$ and $E_{act} \cap E _ {auto} = \emptyset$.
Transitions in $E_{act}$ require explicit activation through an atomic task configuration, whereas transitions in $E_{auto}$ do not require such activation.
Fig.~\ref{HA} illustrates the hybrid automaton used by the truck component in the motivating example in Section~\ref{sec:motivating-example}.
The controllable \textit{notify} transition belongs to $E_{act}$ and sets the target parameters, while the \textit{arrive} transition belongs to $ E _ {auto}$ and triggers when the truck reaches the target region.

\end{definition}

The overall mission may be complex and can be decomposed into various subtasks.
Subtasks may be executed either concurrently or sequentially.
We use a specified structure called \textit{Scenario Task}~\cite{wang_sniff_2024} to describe the whole task plan. 
\begin{definition}
\textbf{(Scenario Task)}.
A scenario task is recursively defined $ST=AT\;|\;(ST \seq ST) \;|\; \allowbreak (ST \parallel ST)$, where $AT$ is an atomic task, $(ST \seq ST)$ denotes the sequential execution of two scenario tasks, and $(ST \parallel ST)$ denotes the parallel composition of two scenario tasks.
Therefore, it can either be a single atomic task or a sequential/parallel compositional task. 
\end{definition}

\begin{definition}
\label{def:atomic-task}
\textbf{(Atomic Task)}.
An atomic task is the smallest task unit, and it specifies which components participate and which action they should choose.
Formally, an atomic task is a tuple
\[
AT = (A_a, U_a, q_a^0, q_a^f),
\]
where $A_a$ is the set of involved component instances, $U_a$ is the set of activated transitions with parameter bindings, and $q_a^0, q_a^f$ specify the expected initial and final modes of the participating components.
Among them, $U_a$ specifies which transitions should be activated, along with their reset parameters.
Figure~\ref{HA}(b) illustrates an example atomic task configuration.
\end{definition}

Signal Temporal Logic (STL) is commonly used to specify requirements for control synthesis in CPSs.
It can specify not only logical relationships but also dense-time constraints.
We use it to capture temporal behaviors and requirements in multi-component CPSs.

\begin{definition}
\textbf{(Signal Temporal Logic~\cite{maler_stl_2004})}.
The syntax of an STL formula $\varphi$ interpreted over a set of atomic predicates $\Pi$ is defined as:
\[
\varphi := \mathit{true} \mid p \mid \neg\varphi_1 \mid \varphi_1 \vee \varphi_2 \mid \varphi_1 \mathcal{U}_{\mathbf{I}}\;\varphi_2.
\]

where $p \in \Pi$, $\mathcal{U}$ denotes the until operator, and $\mathbf{I} \subseteq \mathbb{R}_{\ge 0}$ denotes a timing interval, which can be omitted when it is $[0, \infty)$.
Other standard operators can be derived as follows: $\mathit{false} \equiv \neg\mathit{true}$, $\varphi_1 \wedge \varphi_2 \equiv \neg(\neg\varphi_1 \vee \neg\varphi_2)$, $\varphi_1 \rightarrow \varphi_2 \equiv \neg\varphi_1 \vee \varphi_2$, $\Diamond_I\varphi \equiv \mathit{true}\;\mathcal{U}_{\mathbf{I}}\;\varphi$, and $\Box_I\varphi \equiv \neg\Diamond_I(\neg\varphi)$.
\end{definition}

\textbf{Semantics: } An STL formula $\varphi$ has Boolean semantics that define its satisfaction relation over a signal $s$ at a time $t$.
This satisfaction relation is defined inductively as follows:
\begin{equation}
\begin{split}
    (s, t) \models p & \leftrightarrow \pi_p(s)[t] = \text{True} \\
    (s, t) \models \neg\varphi & \leftrightarrow (s, t) \not\models \varphi \\
    (s, t) \models \varphi_1 \vee \varphi_2 & \leftrightarrow (s, t) \models \varphi_1 \text{ or } (s, t) \models \varphi_2 \\
    (s, t) \models \varphi_1 \mathcal{U}_{[a,b]}\varphi_2 & \leftrightarrow \exists t' \in [t+a, t+b] \text{ s.t. } (s, t') \models \varphi_2 \\
    &\quad \text{ and } \forall t'' \in [t, t'), (s, t'') \models \varphi_1
\end{split}
\end{equation}

STL conventionally uses atomic predicates over continuous states, such as numerical conditions on the variables of CPS components.
In this work, we additionally define \textit{task predicates} to express whether a scenario task has begun or finished.
Thus, our STL formulas may refer to both component states and task status.

\begin{definition}
\label{def:task-predicates}
\textbf{(Task Predicates)}.
For a scenario task $ST$, the predicate $\_TaskBegin(ST)$ indicates that $ST$ has begun, and $\_TaskFinish(ST)$ indicates that $ST$ has finished. 
\end{definition}

\textbf{Semantics: } These predicates have cumulative semantics.
Once the task $ST$ has begun,\linebreak[4] $\_TaskBegin(ST)$ remains true at all later times; otherwise, it is false.
Similarly, once task $ST$ has finished, $\_TaskFinish(ST)$ remains true at all later times; otherwise, it is false.

\begin{definition} 
\label{def:state-predicates}
\textbf{(State Predicates)}.
The state predicate is a numeric condition over the continuous variables of CPS components, such as $\theta(\mathbf{x}) \geq d$.
\end{definition}

Overall, STL formulas over task status predicates specify requirements on task progression, while formulas over state predicates specify requirements on physical trajectories.

\subsection{Motivating Example}
\label{sec:motivating-example}

\begin{figure}[!t]
  \centering
  \vspace{-0.3cm}
  \includegraphics[width=0.9\textwidth]{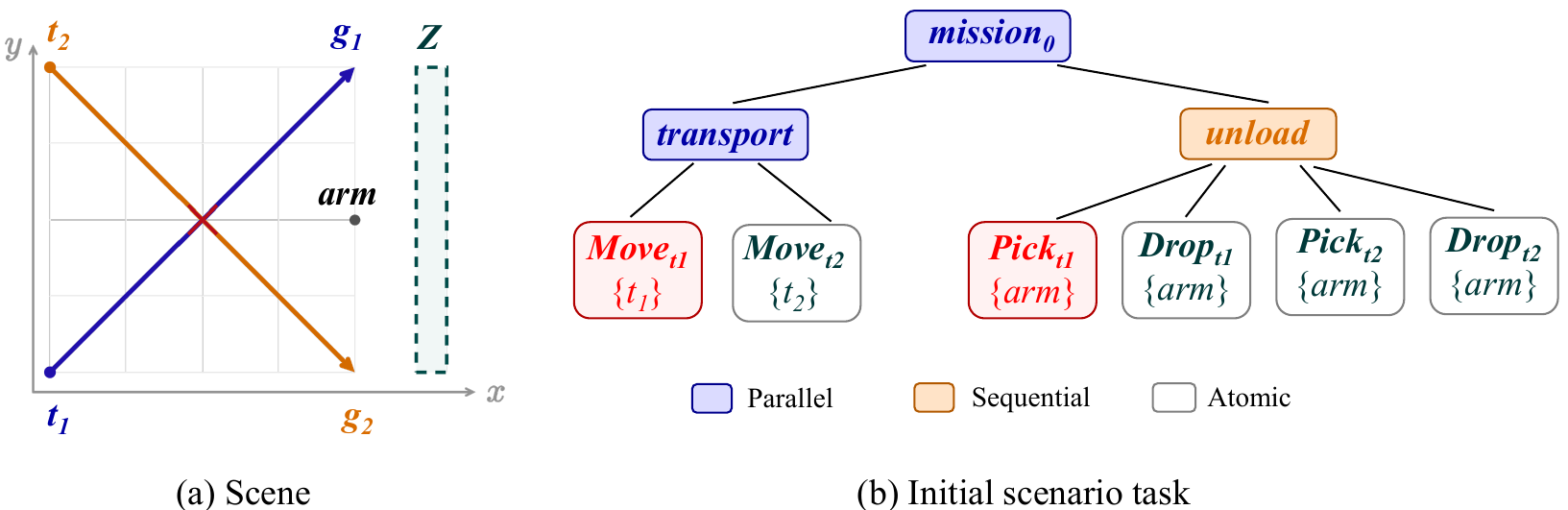}
  \vspace{-0.3cm}
  \caption{Motivating Example. }
  \label{fig:motivating-example}
  \Description{Motivating Example. }
  \vspace{-0.45cm}
\end{figure}

Consider a multi-component CPS scenario with two trucks ($t_1$, $t_2$) and one robotic arm ($arm$).
The system undertakes the following mission: both trucks move to their respective destinations in parallel, arrive within 20 seconds, and maintain a distance of at least 1 metre throughout the mission.
After both trucks arrive, use the arm to unload the object from truck 1 into the place zone, then unload the object from truck 2.
As shown in Fig.~\ref{fig:motivating-example}(a), $t_1$ starts at $(0,0,0)$ and targets $g_1=(10,10,0)$, while $t_2$ starts at $(0,10,0)$ and targets $g_2=(10,0,0)$.
Both trucks travel at $1\,\mathrm{m/s}$.
The arm is based at $(10,5,3)$ and places the objects in $Z=[12,13]\times[0,10]\times\{0\}$.

We use this mission to illustrate both the concepts introduced in Section~\ref{sec:preliminaries} and the two challenges identified in Section~\ref{sec:intro}.
We show an LLM-generated initial task plan in Fig.~\ref{fig:motivating-example}(b), which can be represented as a scenario task:
\begin{equation}
\label{eq:initial-scenario}
\begin{aligned}
\mathit{mission}_0
    &= \mathit{transport}\parallel\mathit{unload},\\
\mathit{transport}
    &= \mathit{move_{t_1}}\parallel\mathit{move_{t_2}}, \\
\mathit{unload}
    &= \mathit{pick_{t_1}}\seq\mathit{drop_{t_1}}\seq
       \mathit{pick_{t_2}}\seq\mathit{drop_{t_2}}.
\end{aligned}
\end{equation}

Its root runs the transportation task \textit{transport} in parallel with the unloading task \textit{unload}.
Transportation consists of two parallel routes, while unloading consists of four sequential actions.
The subtasks $move_{t_1}$ and $move_{t_2}$ move the respective trucks directly to $g_1$ and $g_2$.
The arm executes $pick_{t_1}$ to retrieve the object from $t_1$ at $g_1$, followed by $drop_{t_1}$ to deposit it in $Z$, and repeats the same action for $t_2$.
Thus, the leaf assignments are $\{t_1\}$ for $move_{t_1}$, $\{t_2\}$ for $move_{t_2}$, and $\{\mathit{arm}\}$ for each manipulation task.

Figure~\ref{HA}(b) expands one atomic task, $AT_{move_{t_1}}=(A_a,U_a,q_a^0,q_a^f)$.
Its component set is $A_a=\{t_1\}$; $U_a$ activates $t_1.{notify}$ with target parameters $(10,10,0)$; and $q_a^0=q_a^f=Idle$.
These fields connect the atomic task to the truck HA in Fig.~\ref{HA}: the controllable \textit{notify} transition enters \textit{Moving}, the continuous flow advances the truck toward its target, and the automatic \textit{arrive} transition returns it to \textit{Idle} when the arrival guard holds.
The configuration therefore specifies both the participating component and the behavior it must execute.

For task predicates, two representative LLM-generated constraints are
\begin{align}
\varphi_{\mathrm{order}} &=
\Box\bigl(
\_TaskBegin(\mathit{pick_{t_1}})
\rightarrow
\_TaskFinish(\mathit{move_{t_1}})
\bigr),
\label{eq:example-order}\\
\varphi_{\mathrm{time}} &=
\Diamond_{[0,20]}\_TaskFinish(\mathit{transport}).
\label{eq:example-time}
\end{align}
The first requires truck $t_1$ to have arrived by the time the arm begins unloading the object from $t_1$.
The second requires the transportation task to finish within 20 seconds. 

For state predicates, let $\mathbf{p}_i$ be the position of truck $t_i$.
Two representative constraints are
\begin{align}
\psi_{\mathrm{safe}} &=
\Box\bigl(\|\mathbf{p}_1-\mathbf{p}_2\|_2\geq 1\bigr),
\label{eq:example-safety}\\
\psi_{\mathrm{arrive}} &=
\Diamond_{[0,20]}\Bigl(
\|\mathbf{p}_1-g_1\|_\infty<\epsilon\ \wedge
\|\mathbf{p}_2-g_2\|_\infty<\epsilon\Bigr),
\label{eq:example-arrival}
\end{align}

where $\epsilon$ is a small positive tolerance, and the infinity norm requires the absolute error in each coordinate to be less than $\epsilon$.
The first constraint requires the trucks to maintain a separation of at least 1 metre throughout the mission.
The second requires both trucks to be within their destinations within 20 seconds.

In this example, the LLM initially generates an incorrect plan that executes $\mathit{transport}$ and $\mathit{unload}$ in parallel.
As a result, $\mathit{pick}_{t_1}$ begins when transportation begins, violating constraint~\ref{eq:example-order}.
This inconsistency follows from the task structure and can be detected without analyzing continuous motion.
We show later how this inconsistency is detected to guide revision in Section~\ref{sec:taskallocation}.

Moreover, changing the root composition to $\mathit{transport}\seq\mathit{unload}$ resolves this ordering conflict, but does not by itself ensure physical feasibility.
Under the motion settings stated by LLMs, the trucks start their direct movements together and reach the intersection of their routes at the same time, violating constraint~\ref{eq:example-safety}.
We revisit this example in Section~\ref{sec:repair} to illustrate how this diagnostic information guides revision.
\section{Problem Formulation}
The preceding example illustrates a mission involving multiple CPS components and constraints on their execution.
We now formalize the problem of generating a feasible execution for such a mission.

\noindent\textbf{Inputs.}
An input scene instance is represented as
\begin{equation}
\mathcal{S}=(\mathcal{M},\mathcal{W},\mathcal{H}),
\end{equation}
where $\mathcal{M}$ is the mission expressed in natural language, $\mathcal{W}$ is a structured description of the task environment, and $\mathcal{H}$ contains the functional descriptions and hybrid automaton models of the CPS components.
Among them, the task environment is represented as
\begin{equation}
\mathcal{W}=(\mathcal{Z},\mathcal{O},\mathcal{C'}),
\end{equation}

where $\mathcal{Z}$ is the set of relevant zones, $\mathcal{O}$ describes the objects and obstacles in the scene, and $\mathcal{C'}$ is the set of available component instances.
These elements are provided as structured entries specifying the relevant names and types, together with spatial coordinates or ranges.

\noindent\textbf{Output.}
The desired output is a feasible execution trace $\rho$ over the composed components.
We represent the trace as
\[
\rho =
(\mathbf{q}_0,\mathbf{x}_0,t_0)
\xrightarrow{\mathbf{e}_0}
(\mathbf{q}_1,\mathbf{x}_1,t_1)
\xrightarrow{\mathbf{e}_1}
\cdots ,
\]
where \(\mathbf{q}_i=(q_i^c)_{c\in \mathcal{C}_i}\) denotes the joint discrete state of the active components at time \(t_i\), and \(\mathbf{x}_i=(x_i^c)_{c\in \mathcal{C}_i}\) collects their continuous states (variables).
Here, $\mathcal{C}_i$ is the set of active components, and $\mathbf{e}_i$ denotes the component transitions taken in the corresponding discrete step.
The trace also includes the continuous evolution between successive recorded states. 

The trace must be consistent with the component models in $\mathcal{H}$ and the task environment $\mathcal{W}$, and fulfill the original natural language mission $\mathcal{M}$.

\section{Methodology}

\begin{figure*}[!t]
  \centering
  \vspace{-0.3cm}
  \includegraphics[width=1\textwidth]{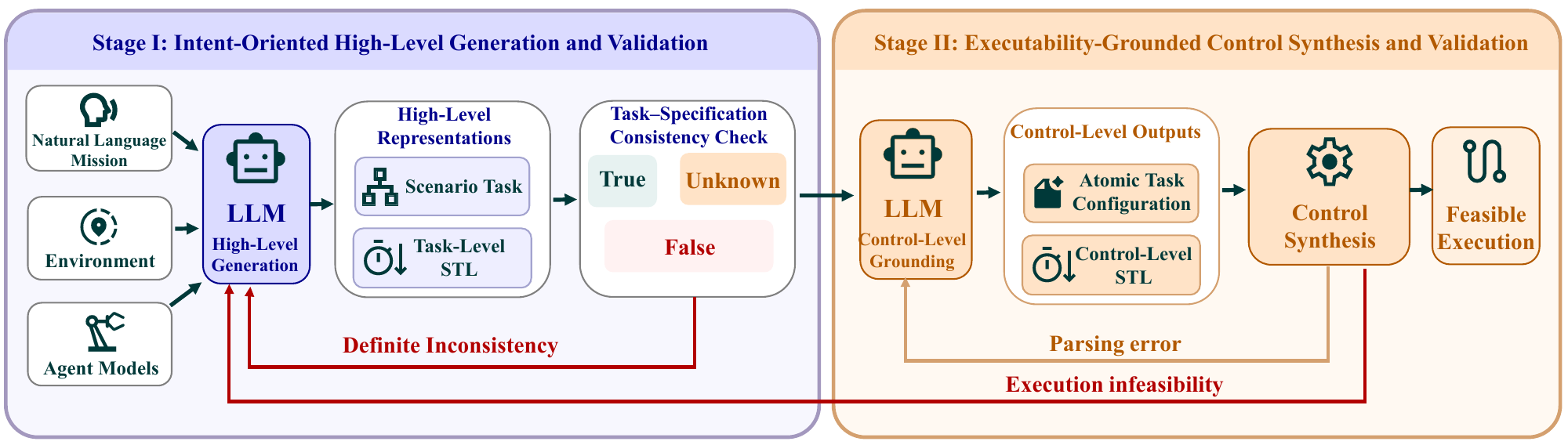}
  \vspace{-0.3cm}
  \caption{Overview of the two-stage framework with hierarchical feedback.}
  \label{framework}
  \Description{framework}
  \vspace{-0.4cm}
\end{figure*}

\subsection{Framework Overview}
\label{sec:framework}

To solve this problem, our framework generates a formal mission description:
\begin{equation}
\label{eq:controller-input}
\mathcal{I}=(T,\Phi).
\end{equation}
Here, $T$ is a scenario task that organizes the mission into subtasks through sequential and parallel composition, and $\Phi$ contains STL formulas specifying the objectives and constraints for control synthesis.
A feasible trace synthesized from $\mathcal{I}$ must follow $T$ and satisfy $\Phi$.

The two challenges discussed in Section~\ref{sec:intro} arise in constructing this description: the semantic gap concerns whether $\mathcal{I}$ captures $\mathcal{M}$, whereas the feasibility gap concerns whether a feasible execution can be synthesized from $\mathcal{I}$.
We address these challenges through a framework with two stages and hierarchical feedback, as shown in Fig.~\ref{framework}.

\begin{algorithm}[!t]
\DontPrintSemicolon
\footnotesize
\SetKwInOut{Input}{Input}
\SetKwInOut{Output}{Output}
\SetKwFunction{CheckSyntax}{CheckSyntax}
\SetKwFunction{CheckConsistency}{CheckConsistency}
\SetKwFunction{InvokeControlSynthesizer}{InvokeControlSynthesizer}
\SetKwFunction{ToolFailureReport}{ToolFailureReport}
\SetKw{ContinueLoop}{continue}
\SetKw{BreakLoop}{break}

\newcommand{\AlgOneCommentStyle}[1]{\textcolor{blue}{\normalfont #1}}
\SetCommentSty{AlgOneCommentStyle}
\SetKwComment{tcp}{\(\triangleright\)\ }{}

\Input{Scene instance $\mathcal{S}$, iteration budgets $K_1,K_2,K_3$}
\Output{Generated instance $\mathcal{I}$ and feasible execution trace $\rho$, or failure report}
\caption{Two-Stage Synthesis with Hierarchical Feedback}
\label{alg:framework}

\For{$i \leftarrow 1$ \KwTo $K_1$}{
    \For{$j \leftarrow 1$ \KwTo $K_2$}{
        Prompt the LLM to generate scenario task $T$ and task-level STLs $\Phi_T$ from $\mathcal{S}$\;
        $R_T \leftarrow \CheckSyntax(\Phi_T)$\;
        \If{$R_T$ reports no syntax error}{
            $R_T \leftarrow \CheckConsistency(T,\Phi_T)$
            \tcp*[r]{see Algorithm~\ref{alg:stlconsistency}}
        }
        \If{$R_T$ reports no task-level error}{
            \BreakLoop\;
        }
        Update the stage-I prompt with $R_T$
        \tcp*[r]{consistency-guided revision}
    }

    \For{$\ell \leftarrow 1$ \KwTo $K_3$}{
        Prompt the LLM to generate atomic task configurations for each $AT$ and control-level STLs $\Phi_C$\;
        $R_C \leftarrow \CheckSyntax(T,\Phi_C)$\;
        \If{$R_C$ reports a control-level parsing error}{
            Update the stage-II prompt with $R_C$
            \ContinueLoop\;
        }
        $\Phi \leftarrow \Phi_T \cup \Phi_C$\;
        $\rho \leftarrow \InvokeControlSynthesizer(T,\Phi)$\;
        \If{$\rho$ is feasible}{
            $\mathcal{I} \leftarrow (T,\Phi)$\;
            \Return $\mathcal{I},\rho$\;
        }
        $R \leftarrow \ToolFailureReport()$
        \tcp*[r]{localized execution diagnostics}
        Update the stage-I prompt with $R$
        \tcp*[r]{diagnostic-guided replanning}
        \BreakLoop\;
    }
}
\Return failure report\;
\end{algorithm}

Given a scene instance $\mathcal{S}$, we prompt an LLM to decompose the whole objective and construct the hierarchical structure of $T$ through sequential and parallel composition.
In this stage, each atomic task initially contains a natural language description, stating its subgoal, participating components, and relevant objects or regions.
For example, the task $\mathit{move}_{t_1}$ in the motivating example can be described as ``move truck $t_1$ to $g_1=(10,10,0)$.''
This stage also generates \emph{task-level STL specifications} $\Phi_T$, which are formulas over the task predicates introduced in Definition~\ref{def:task-predicates}.
Before proceeding, the framework checks the syntax in $\Phi_T$ and uses a novel algorithm to check the inconsistencies between the plan and these specifications.
Detected errors are returned to Stage~I to guide revision of the task plan or the specifications.

Once the first stage reports no error, we prompt an LLM to convert the natural language description of each atomic task into the structured representation defined in Definition~\ref{def:atomic-task}.
Using the environment description $\mathcal{W}$ and component models $\mathcal{H}$, this stage specifies the participating components, activated transitions, and expected initial and final states.
This stage also generates \emph{control-level STL specifications} $\Phi_C$ over the state predicates introduced in Definition~\ref{def:state-predicates}.
Representative constraints can be found in Section~\ref{sec:motivating-example}.
Once all atomic task configurations have been generated, the completed $T$ and the combined specifications $\Phi=\Phi_T\cup\Phi_C$ form the tool input $\mathcal{I}$.
The framework checks the syntax of $\Phi_C$ and task configurations.
Errors detected are directly returned to Stage~II for correction without revising the accepted task.
Once the outputs pass these checks, the tool attempts to synthesize a feasible execution trace $\rho$ from $\mathcal{I}$ and the scene information.

If control synthesis fails, the diagnostic reports are returned to Stage~I as replanning feedback.
They guide revisions to the hierarchical structure of the task.
This feedback structure distinguishes local syntax corrections from replanning informed by control synthesis failures.
The framework returns the first feasible trace satisfying the generated specifications, or reports failure when the iteration budgets are exhausted.
Algorithm~\ref{alg:framework} summarizes the process.

\subsection{Stage I: Intent-Oriented Task Generation and Consistency Checking}
\label{sec:taskallocation}

In this stage, the LLMs are instructed to generate a task plan and a set of task-level specifications, as mentioned before.
Generating both representations makes these requirements explicit and allows checking conflicts between them before control synthesis.

The main difficulty is that $T$ determines only a partial order over task events, rather than a complete timed execution.
The checker must therefore reason about concurrent tasks without fixing their execution times or imposing an arbitrary order.
We develop an algorithm that evaluates $\Phi_T$ over this partial order using three-valued logic.

\subsubsection{Consistency Checking Algorithm}

\begin{algorithm}[!tbp]
\DontPrintSemicolon
\footnotesize
\SetKwInOut{Input}{Input}
\SetKwInOut{Output}{Output}
\Input{Task tree $T$; STL formula $\varphi$}
\Output{Value of $\varphi$ at the entry event}
\caption{Task Plan and STL Consistency Checking}
\label{alg:stlconsistency}

Construct $G_D$ and $H$ from $T$; let $z_{e_0}$ be the entry node\;

\ForEach{subformula $\psi$ of $\varphi$ with operands processed before their parent}{
    \uIf{$\psi$ is a Boolean constant}{
        Set $V_z(\psi)$ to that constant for every $z$\;
    }
    \uElseIf{$\psi$ is a task predicate}{
        Compute $V_z(\psi)$ using Eq.~\eqref{eq:task-predicate-values}\;
    }
    \uElseIf{$\psi$ is a Boolean combination}{
        Combine its operand values at each node using Kleene logic\;
    }
    \uElseIf{$\psi$ is an unbounded temporal formula}{
        \ForEach{$z$ in reverse topological order of $H$}{
            \uIf{$\operatorname{Succ}_H(z)=\emptyset$}{
                Use $V_z(\alpha)$ for $\Diamond\alpha$ or $\Box\alpha$,
                and $V_z(\beta)$ for $\alpha\mathcal{U}\beta$\;
            }
            \Else{
                Compute $S_z(\psi)$ by Eq.~\eqref{eq:successor-value}\;
                Compute $V_z(\psi)$ by Eq.~\eqref{eq:temporal-values}\;
            }
        }
    }
\Else{
        Set $V_z(\psi)\leftarrow\mathsf{Unknown}$ for every $z$\;
    }
}
\Return $V_{z_{e_0}}(\varphi)$\;
\end{algorithm}

For each task (node) $ST$ in $T$, we introduce its start event $ST.b$ and finish event $ST.f$.
These events determine when $\_TaskBegin(ST)$ and $\_TaskFinish(ST)$ become true.
We assume that every task has positive duration, sequential subtasks start when their predecessors finish, and parallel subtasks share a common start event.
Based on these relations, the algorithm first constructs an event relation graph $G=(V, E_=, E_<)$ where $V$ contains these events, $E_=$ records simultaneous events, and $E_<$ records strict precedence (one event must occur strictly before another).
The graph $G$ captures the logical dependencies among events and is constructed directly from the task plan, as follows. 

Every task contributes $ST.b<ST.f$.
A sequential task $ST=(st_1 \seq st_2 \seq \cdots \seq st_k)$ contributes
\begin{equation}
\label{eq:sequence-events}
ST.b=st_1.b,\qquad
st_i.f=st_{i+1}.b\ (1\leq i<k),\qquad
st_k.f=ST.f.
\end{equation}
A parallel task $ST=(st_1\parallel st_2\parallel\cdots\parallel st_k)$ contributes
\begin{equation}
 ST.b = st_i.b \quad (1\leq i\leq k).
\end{equation}

The algorithm does not impose ordering between $st_i.f$.
Although parallel subtasks share a common start event, they may finish at different times.
However, if the parallel task has a subsequent serial successor $s$, the algorithm adds $y.f<s.f$ for every task $y$ in the parallel subtree.
This follows from the sequential ordering between the parallel task and $s$.

After constructing $G$, the algorithm merges simultaneous events in $E_=$ using union-find and projects the strict edges in $E_<$ onto their representatives, obtaining the directed acyclic graph (DAG) $G_D=(V_D, E_D)$.
We write $\operatorname{rep}(e)\in V_D$ for the representative of each event $e\in V$, and write $u\preceq v$ when $u=v$ or $u\prec v$.
To account for the time between ordered events, the algorithm constructs a new evaluation graph $H$ from $G_D$.
Each representative $e$ becomes an event node $z_e$, and each edge $u\to v \in E_D$ is replaced by $z_u\to z_{uv}\to z_v$.
The interval node $z_{uv}$ represents the open interval between $u$ and $v$, during which other task events may occur.

Finally, the algorithm assigns each subformula a value in $\{\mathsf{True},\mathsf{False},\mathsf{Unknown}\}$ at every node of $H$.
Here, $\mathsf{Unknown}$ means that the event ordering alone is insufficient to decide the value.

\noindent\textbf{Evaluating predicates.}
For a task predicate $p$, let $\operatorname{evt}(p)$ be the representative of its associated start or finish event.
Its value is
\begin{equation}
\label{eq:task-predicate-values}
\begin{aligned}
V_{z_e}(p)&=
\begin{cases}
\mathsf{True}, & \operatorname{evt}(p)\preceq e,\\
\mathsf{False}, & e\prec\operatorname{evt}(p),\\
\mathsf{Unknown}, & \text{otherwise},
\end{cases}
&
V_{z_{uv}}(p)&=
\begin{cases}
\mathsf{True}, & \operatorname{evt}(p)\preceq u,\\
\mathsf{False}, & v\preceq\operatorname{evt}(p),\\
\mathsf{Unknown}, & \text{otherwise}.
\end{cases}
\end{aligned}
\end{equation}
These rules determine whether the event has already occurred or must occur later.
For an interval node, a definite value holds throughout the represented interval.
State predicates are assigned $\mathsf{Unknown}$ because their values require continuous states. 

\noindent\textbf{Evaluating Boolean connectives.}
Subformulas are evaluated from the leaves of the formula syntax tree upward.
We evaluate Boolean connectives using Kleene three-valued logic.
For example, $\mathsf{False}\land\mathsf{Unknown}=\mathsf{False}$, whereas $\mathsf{True}\lor\mathsf{Unknown}=\mathsf{True}$.

\noindent\textbf{Evaluating temporal formulas.}
For an unbounded temporal formula, the algorithm traverses $H$ in reverse topological order.
For a node $z$ with successors, let
\begin{equation}
\label{eq:successor-value}
S_z(\psi)=
\bigwedge_{z'\in\operatorname{Succ}_H(z)} V_{z'}(\psi).
\end{equation}
The conjunction returns $\mathsf{False}$ if any successor is false, $\mathsf{True}$ if all are true, and $\mathsf{Unknown}$ otherwise.
The temporal recurrences are
\begin{equation}
\label{eq:temporal-values}
\begin{aligned}
V_z(\Diamond\alpha)
 &=V_z(\alpha)\lor S_z(\Diamond\alpha),\\
V_z(\Box\alpha)
 &=V_z(\alpha)\land S_z(\Box\alpha),\\
V_z(\alpha\mathcal{U}\beta)
 &=V_z(\beta)\lor
   \bigl(V_z(\alpha)\land S_z(\alpha\mathcal{U}\beta)\bigr).
\end{aligned}
\end{equation}
At a node with no successors, the respective base values are $V_z(\alpha)$, $V_z(\alpha)$, and $V_z(\beta)$.
Operators with interval $[0,0]$ reduce to their current-time operands: $\Diamond_{[0,0]}\alpha$ and $\Box_{[0,0]}\alpha$ to $\alpha$, and $\alpha\mathcal{U}_{[0,0]}\beta$ to $\beta$.
Other bounded temporal operators are assigned $\mathsf{Unknown}$, since the graph does not determine elapsed time.

Algorithm~\ref{alg:stlconsistency} summarizes the check process.
It returns the value at $z_{e_0}$, where $e_0=\operatorname{rep}(T.b)$.
After constructing the event graphs, the evaluation described above takes $O(|\varphi|(n+m))$ time, where $|\varphi|$ is the number of subformulas, and $n$ and $m$ are the numbers of nodes and edges in $H$, respectively. 

\begin{figure*}[!t]
  \centering
  \vspace{-0.3cm}
  \includegraphics[width=0.95\textwidth]{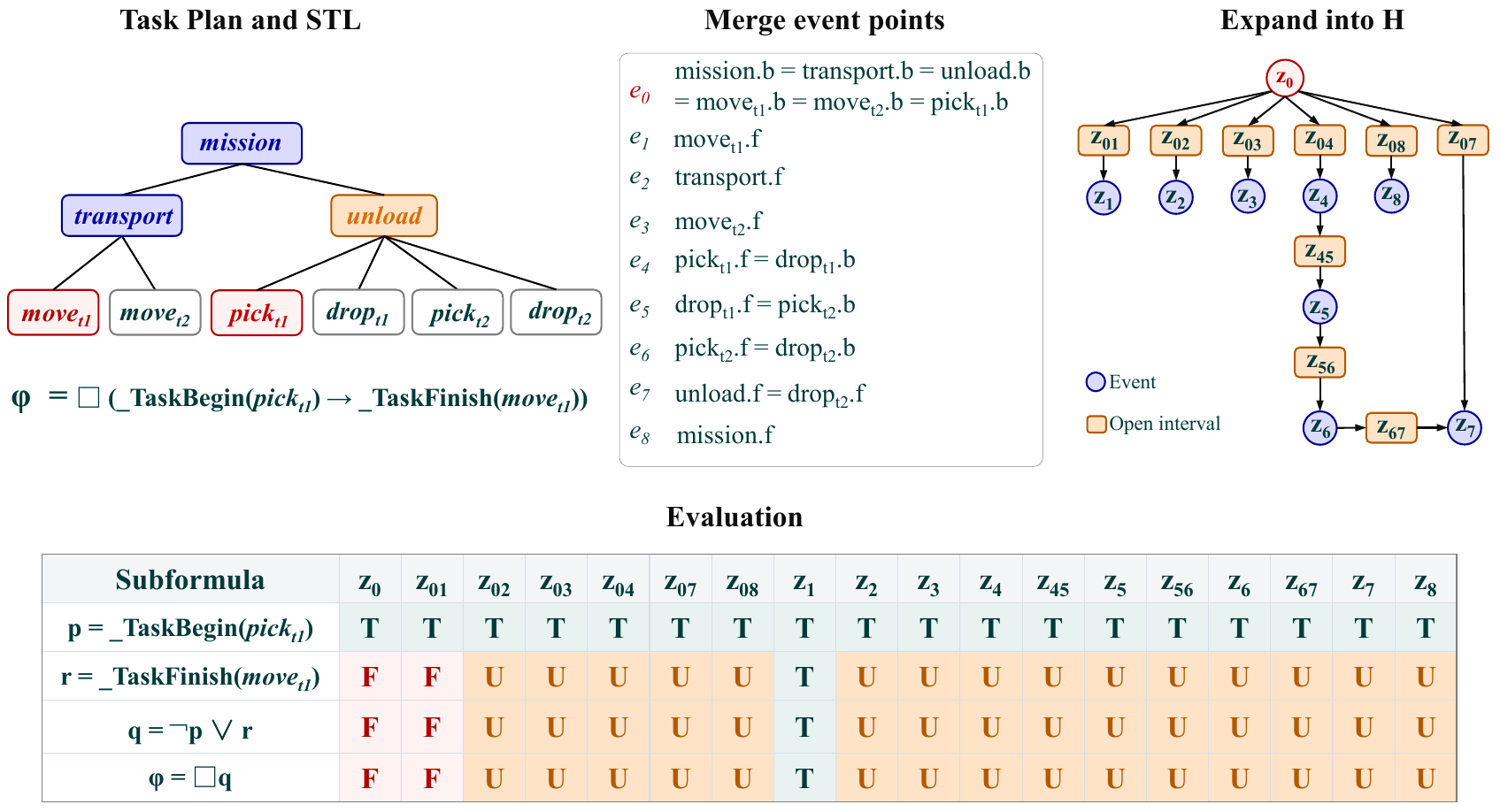}
  \vspace{-0.2cm}
  \caption{Task plan and STL consistency checking for the motivating example.}
  \label{fig:alg}
  \Description{Consistency Checking for Motivating Example.}
  \vspace{-0.5cm}
\end{figure*}

\noindent\textbf{Proposition 1 (Soundness of the consistency checking algorithm).}
Let $\mathcal{R}(G_D)$ denote the set of all event-time assignments $\tau:V_D\rightarrow\mathbb{R}_{\geq 0}$ satisfying $\tau(u)<\tau(v)$ for every strict edge $(u,v)\in E_D$.
Each $\tau$ induces a cumulative Boolean signal $s_\tau$ over the task predicates,where, for a task predicate $p_e$ associated with event $e$,
\begin{equation}
(s_\tau,t)\models p_e
\quad\Longleftrightarrow\quad
\tau(e)\leq t.
\end{equation}
The algorithm satisfies
\begin{equation}
V_{z_{e_0}}(\varphi)=\mathsf{False}
\;\Longrightarrow\;
\forall \tau\in\mathcal{R}(G_D),\;
(s_\tau,\tau(e_0))\not\models\varphi,
\label{eq:consistency-soundness}
\end{equation}
where $e_0=\operatorname{rep}(T.b)$ is the entry event and $z_{e_0}$ is its corresponding event node in $H$.
Therefore, every formula rejected by the algorithm is genuinely inconsistent with every timing realization permitted by the event-order abstraction.

The result follows by mutual structural induction on definite truth and falsity, with a nested induction over the reverse topological order of $H$ for temporal formulas.
Equation~\eqref{eq:task-predicate-values} assigns a definite value to a task predicate only when the partial order guarantees that its associated event has already occurred, or has not yet occurred, throughout the time region represented by the current node.
Kleene three-valued Boolean operations preserve the induction invariant.

For temporal formulas, the successor aggregation in Eq.~\eqref{eq:successor-value} and the recurrences in Eq.~\eqref{eq:temporal-values} preserve the same invariant.
In particular, every direct edge $u\to v$ in $G_D$ is represented in $H$ by an interval node corresponding to the entire open interval $(\tau(u),\tau(v))$.
If $u$ has multiple direct successors, their outgoing intervals share a nonempty time interval immediately after $\tau(u)$, because all outgoing edges are strict and the graph is finite.
Consequently, a false successor certifies a violation over an actual time interval rather than merely along one graph path.
This prevents an intermediate event from being skipped and producing a spurious inconsistency report.

\subsubsection{Feedback and Revision}

Syntax errors and formulas evaluated as $\mathsf{False}$ are returned to the same LLM session in Stage~I to revise $T$, $\Phi_T$, or both.
In the motivating example,  $\mathit{pick}_{t_1}.b$ maps to the entry event $e_0$, while $e_0\prec\operatorname{rep}(\mathit{move}_{t_1}.f)$.
In Eq.~\eqref{eq:example-order}, the algorithm returns $\mathsf{False}$ for $\varphi_{\mathrm{order}}$, prompting a revision such as replacing the root composition with $\mathit{transport}\seq\mathit{unload}$.
Therefore, the antecedent of the implication is true and its consequent is false at the entry event.
Figure~\ref{fig:alg} shows the concrete process of evaluating the motivating example in Section~\ref{sec:motivating-example}.

\subsection{Stage II: Executability-Oriented Control Synthesis and Diagnostic Feedback}
\label{sec:controlgeneration}
\label{sec:repair}

After task-level checking, feasibility still depends on how the atomic tasks are realized by the CPS components.
Stage~II therefore connects each task description to concrete component behaviors and submits the resulting mission description to control synthesis.
Using the environment description $\mathcal{W}$ and component models $\mathcal{H}$, we prompt an LLM to convert each task description into the configuration $AT$ defined in Definition~\ref{def:atomic-task}.
A key step is selecting the transitions to activate.
A component HA model may provide several functions, but only a subset is needed for a given task.
For example, a robotic arm may expose transitions for grasping, and placing.
To ground a placing task, the LLM should select the placing transition and binds the target position through reset variables.
The LLM also generates control-level STL specifications $\Phi_C$ as mentioned before.

The framework therefore submit the completed mission description $\mathcal{I}=(T,\Phi)$ to the underlying control synthesis tool~\cite{wang_sniff_2024}.
We extend the tool's diagnostic reports to connect failures in the generated behavior to the tasks and components involved, providing feedback for revising $\mathcal{I}$.

\noindent\textbf{Input validation.}
Before synthesis, an STL parser checks the syntax of $\Phi_C$, while the tool's input parser validates the task configurations in $T$.
Errors detected by these checks are directly returned to Stage~II for correction.
Once the input passes validation, the tool attempts to synthesize a feasible execution under the component models and the specifications.

\noindent\textbf{Control Synthesis diagnostics.}
For failures reported during control synthesis, we extract diagnostic information about the failed requirement or component behavior and associate it with the relevant tasks and components.
The following diagnostic categories support this connection between synthesis failures and task revision.

\emph{STL violations.}
The report identifies the violated specification and the tasks active when the violation is detected. 

\emph{Conflicting component assignments.}
When parallel tasks demand incompatible simultaneous use of the same component, the report identifies that component and the conflicting tasks.
This exposes an allocation conflict that may require changing the participating components or the task composition.

\emph{Invariant violations.}
When a component's continuous evolution violates the invariant $Inv_q$ of its current mode, the report identifies the invariant, the affected component, and the associated active tasks.
This reveals a conflict between the planned behavior and the conditions under which the component may remain in that mode.

\emph{Guard failures.}
When a required transition cannot be taken because its guard $G_e$ is unsatisfied, the report identifies the transition, the unsatisfied guard, and the associated tasks.
The failure concerns a transition needed for task progress.

\noindent\textbf{Feedback routing and revision.}
Synthesis failures are returned to Stage~I because resolving them may require changes to task decomposition or component allocation.
The diagnostic reports connect the observed failure to the relevant tasks and requirements, guiding the LLM in reconsidering the plan rather than merely requesting another attempt.
The revised task plan and task-level specifications undergo the Stage~I checks again before Stage~II regenerates the corresponding configurations and control-level specifications.
This process continues until a feasible trace is obtained or the iteration budgets are exhausted.

\begin{figure}[!t]
  \centering
  \vspace{-0.3cm}
  \includegraphics[width=0.9\textwidth]{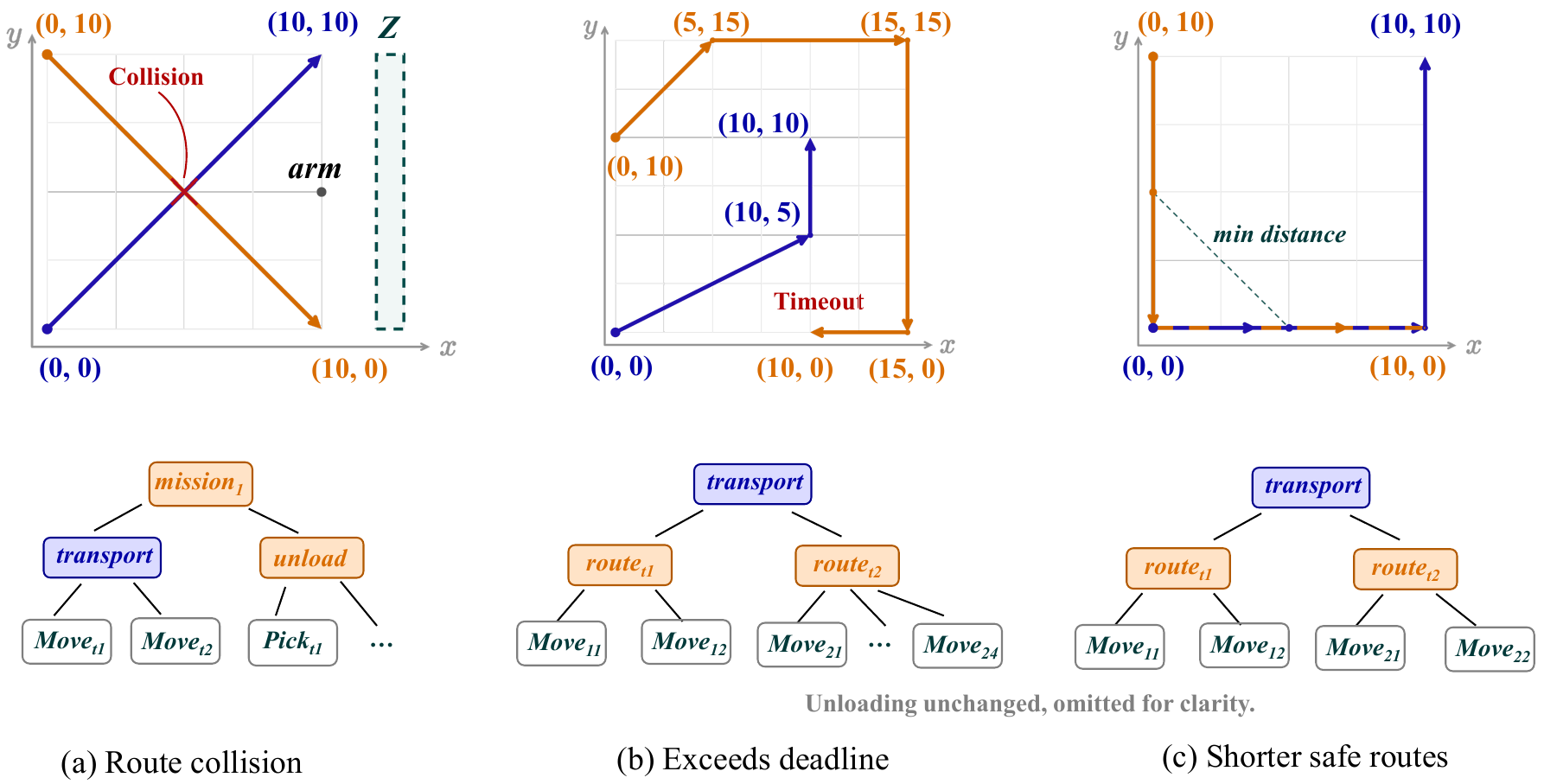}
  \vspace{-0.4cm}
  \caption {Task plan repair guided by control synthesis feedback: resolving collision and deadline violations. }
  \label{fig:example2}
  \Description{Execution Feedback-Guided Repair. }
  \vspace{-0.3cm}
\end{figure}

\noindent\textbf{Example.}
We continue the motivating example after the ordering conflict has been corrected: transportation remains parallel, and unloading starts after transportation finishes, which shows in Fig.~\ref{fig:example2}(a).
With the initial direct routes, the trucks reach $(5,5)$ simultaneously, violating the separation requirement in Eq.~\eqref{eq:example-safety}.
The synthesis tool reports the violated formula together with the two movement tasks, and this feedback is returned to Stage~I.

The first revision replaces each direct movement with a sequence of waypoint movements while retaining parallel transportation, producing Fig.~\ref{fig:example2}(b).
Truck~1 follows $(0,0)\rightarrow(10,5)\rightarrow(10,10)$, and truck~2 follows $(0,10)\rightarrow(5,15)\rightarrow(15,15)\rightarrow(15,0) \rightarrow(10,0)$.
Although the detour avoids the original collision, truck~2 must travel approximately $37\,\mathrm{m}$.
At $1\,\mathrm{m/s}$, it cannot meet the 20-second deadline, so a further revision is required.

After the second feedback, truck~1 follows $(0,0)\rightarrow(10,0)\rightarrow(10,10)$, while truck~2 follows $(0,10)\rightarrow(0,0)\rightarrow(10,0)$.
Each route is $20\,\mathrm{m}$ long, corresponding to 20 seconds of ideal travel (Fig.~\ref{fig:example2}(c).).
The trucks traverse their shared road segment at different times and maintain a minimum separation of approximately $7\,\mathrm{m}$.
In this example, the feedback changes how the mission is carried out while retaining its safety, timing, and parallelism requirements.
It therefore illustrates how control synthesis diagnostics guide revisions to task decisions.

\section{Experiment Evaluation}
\label{sec:exp}
This section evaluates whether our proposed framework can turn scene instances into feasible executions for multi-component CPSs.
We focus on the following three key questions:

\textbf{RQ1:} Can the full framework generate feasible controls that satisfy the original natural language missions across heterogeneous multi-component CPS scenes and increasingly complex objectives?

\textbf{RQ2:} How much does our hierarchical feedback mechanism contribute to the success rate of the control generation?

\textbf{RQ3:} Can our work be extended to other scenarios, such as multi-robot tasks? How does it perform compared with existing LLM-to-planning works?

\subsection{Experimental Setup}
\textbf{Implementation.}
We implement the prototype of our framework in Python for convenient access to LLM interfaces.
The underlying control synthesis tool SNIFF~\cite{wang_sniff_2024} is implemented in C++ for efficient low-level solving.
We use it as the underlying search engine for feasible executions and extend its diagnostic interface to support LLM-guided revision.
Three models, including DeepSeek-V4-Flash~\cite{deepseekai_deepseekv4_2026}, GPT-5 (version gpt-5-2025-08-07), and Qwen3.7-Plus, are involved in the experiments.
DeepSeek and Qwen are used with thinking mode enabled, and GPT-5 is configured with a medium reasoning effort.
We use 3, 3, and 4 as the iteration budgets for \(K_1\), \(K_2\), and \(K_3\), respectively, throughout the experiments.
These settings allow repeated repair attempts while limiting the overall running time.

\textbf{Benchmark.}
The benchmark suite contains seven heterogeneous scenes that we built manually, two of which are adapted from AI2-THOR~\cite{eric_ai2thor_2022} simulation environments.
As summarized in Table~\ref{tab:scenes}, the scenes cover diverse application domains, including object manipulation, spatial reasoning, household tasks, disaster response, and more.
Each scene contains four natural language objectives.
From objectives~1 to objectives~4 (O1--O4), the number of tasks or constraints is progressively increased during construction.
The LLMs may generate different task decompositions and STL formulas for the same objective.
Table~\ref{tab:scenes} therefore reports the average numbers of generated tasks (including both atomic and composite tasks) and STL formulas to describe the sizes of the resulting plans and specifications.

\begin{table*}[!t]
\centering
\footnotesize
\caption{Benchmark scenes and average numbers of generated task nodes and STL formulas.}
\label{tab:scenes}
\setlength{\tabcolsep}{2.5pt}
\renewcommand{\arraystretch}{1}

\begin{tabularx}{\linewidth}{
  @{}
  p{0.06\linewidth}
  >{\raggedright\hyphenpenalty=10000\exhyphenpenalty=10000\arraybackslash}p{0.18\linewidth}
  >{\raggedright\arraybackslash}X
  *{8}{r}
  @{}
}
\toprule
\textbf{Scene} & \textbf{Component types} & \textbf{Objective summary} &
\multicolumn{4}{c}{\textbf{Tasks}} &
\multicolumn{4}{c}{\textbf{STL formulas}} \\
\cmidrule(lr){4-7}\cmidrule(l){8-11}
& & & \textbf{O1} & \textbf{O2} & \textbf{O3} & \textbf{O4} &
\textbf{O1} & \textbf{O2} & \textbf{O3} & \textbf{O4} \\
\midrule
Scene 1 & Robotic arms and mobile carriers &
Stack blocks into towers at specified locations while conforming to constraints. &
16 & 16 & 26 & 30 & 21 & 20 & 27 & 30 \\
Scene 2 & UAVs &
Access different rooms while satisfying complex ordering constraints.
This scene is adapted from SELP~\cite{wu_selp_2025}. &
12 & 11 & 12 & 12 & 24 & 26 & 29 & 31 \\
Scene 3 & Robotic arm, coffee/meal delivery robot, cleaning robot &
Coordinate multiple household tasks, including spill cleaning, breakfast preparation, and food delivery &
10 & 15 & 16 & 21 & 13 & 16 & 15 & 23 \\
Scene 4 & Crane, robot dog, ambulance, UAV, dozer &
Complete rescue during an earthquake, including clearing roads, building bridges, and reaching injured people. &
4 & 11 & 13 & 16 & 8 & 16 & 18 & 19 \\
Scene 5 & Trucks and robot arm &
Move the trucks to their destinations in parallel.
Then, have the arm pick up the objects from the trucks and drop them off. &
5 & 6 & 13 & 16 & 10 & 10 & 16 & 20 \\
Scene 6 & Mobile household agents & Open the toaster, close the light switch, and pick up the tomato while staying within the valid area. &
5 & 4 & 9 & 13 & 20 & 21 & 29 & 37 \\
Scene 7 & Mobile household agents &
Close the light switch and put the object into the coffee machine while staying within the allowed area. &
6 & 5 & 10 & 11 & 22 & 23 & 30 & 29 \\
\bottomrule
\end{tabularx}
\vspace{-10pt}
\end{table*}

\textbf{Ablation study.}
To evaluate the effectiveness of the proposed hierarchical feedback mechanism, we compare four strategies on the same set of scene instances:
\begin{itemize}[leftmargin=1.5em]
    \item \textbf{Full:} our complete method with all feedback channels enabled. 
    \item \textbf{No-TSTL-Check:} disables the syntactic and semantic consistency check for task-level STLs in stage 1 while keeping the rest of the pipeline unchanged.
    \item \textbf{No-Replan:} stops a run when the synthesis tool reports a failure, rather than routing back to stage 1 for replanning. 
    \item \textbf{Simple-Feedback:} preserves the same feedback routing as \textit{Full}, but replaces the detailed reports with a simple retry prompt.
Just inform the LLMs that an error has occurred.
\end{itemize}

To ensure a fair comparison: (1) the run scripts randomize the execution order across strategies, and (2) each scene-objective case is repeated five times to account for the stochasticity of LLMs.
In total, this ablation study evaluates 3 different LLMs across 7 scenes, 4 objectives per scene, 5 repetitions, and 4 strategies, resulting in \textbf{1680} runs.

\textbf{Comparison with multi-robot planning methods.}
We further compare our framework with two representative LLM-based task planning approaches, SMART-LLM~\cite{kannan_smartllm_2024} and NL2HLTL2PLAN~\cite{xu_nl2hltl2plan_2025}, on two AI2-THOR household scenes (scene~6 and scene~7).
Each scene contains four objectives, and similarly, each case is repeated five times for every method. 

The three frameworks differ in their outputs.
Our framework generates control trajectories, which are subsequently converted into executable AI2-THOR Python code within an automated script.
SMART-LLM generates Python code directly.
NL2HLTL2PLAN generates low-level trajectories defined on its own grid abstraction, which must also be converted into code.

\begin{table}[!t]
\centering
\footnotesize
\caption{Generation and verified success rates of \textit{Full} with GPT-5.}
\label{tab:full-gpt5-objectives}
\setlength{\tabcolsep}{1pt}
\renewcommand{\arraystretch}{0.98}

\resizebox{\linewidth}{!}{\begin{tabular}{@{}l*{10}{c}@{}}
\toprule
\multirow{3}{*}{\textbf{Scene}}
& \multicolumn{10}{c}{\textbf{Success rate (\%)}} \\
\cmidrule(l){2-11}
& \multicolumn{2}{c}{\textbf{O1}}
& \multicolumn{2}{c}{\textbf{O2}}
& \multicolumn{2}{c}{\textbf{O3}}
& \multicolumn{2}{c}{\textbf{O4}}
& \multicolumn{2}{c}{\textbf{Total}} \\
\cmidrule(lr){2-3}
\cmidrule(lr){4-5}
\cmidrule(lr){6-7}
\cmidrule(lr){8-9}
\cmidrule(l){10-11}
& \textbf{Generation} & \textbf{Verified}
& \textbf{Generation} & \textbf{Verified}
& \textbf{Generation} & \textbf{Verified}
& \textbf{Generation} & \textbf{Verified}
& \textbf{Generation} & \textbf{Verified} \\
\midrule
Scene 1 & 80  & 80  & 100 & 100 & 60  & 60  & 100 & 100 & 85  & 85  \\
Scene 2 & 100 & 100 & 100 & 100 & 40  & 40  & 80  & 80  & 80  & 80  \\
Scene 3 & 100 & 100 & 100 & 100 & 100 & 60  & 100 & 60  & 100 & 80  \\
Scene 4 & 100 & 100 & 100 & 80  & 80  & 80  & 100 & 100 & 95  & 90  \\
Scene 5 & 100 & 100 & 100 & 100 & 100 & 100 & 40  & 40  & 85  & 85  \\
Scene 6 & 100 & 100 & 100 & 100 & 100 & 100 & 80  & 80  & 95  & 95  \\
Scene 7 & 100 & 100 & 100 & 100 & 100 & 100 & 100 & 100 & 100 & 100 \\
\midrule
\textbf{Average}
& \textbf{97.1}  & \textbf{97.1}
& \textbf{100.0} & \textbf{97.1}
& \textbf{82.9}  & \textbf{77.1}
& \textbf{85.7}  & \textbf{80.0}
& \textbf{91.4}  & \textbf{87.9} \\
\bottomrule
\end{tabular}}
\end{table}

\subsection{Success Rates Across Heterogeneous Scenes}

To answer RQ1, we examine not only whether the framework can generate feasible controls, but also whether these controls satisfy the intent of the original natural language missions.
Table~\ref{tab:full-gpt5-objectives} reports the results of \textit{Full} with GPT-5, which achieves the highest overall success rate among the three evaluated models.

We distinguish two success criteria.
\emph{Generation success} means that the control synthesis tool successfully generates a feasible control from the LLM-generated task plan and specifications.
\emph{Verified success} further requires the generated control to satisfy the original natural language mission, as determined by automated checking scripts together with manual inspection.
The latter therefore provides a stricter end-to-end measure of whether the framework translates user intent into a feasible and semantically correct control.
Unless otherwise stated, all success rates reported for our framework from RQ2 onward use this the \textit{Verified success} criterion.

Table~\ref{tab:full-gpt5-objectives} shows that \textit{Full} achieves an overall generation success rate of 91.4\% and a verified success rate of 87.9\%, a gap of only 3.5\%.
This small gap indicates that, in most cases where control synthesis succeeds, the resulting control also satisfies the requirements expressed in the original natural language mission.

The results remain strong as the objectives become more challenging.
From O1 to O4, the objectives introduce additional tasks or constraints, leading to larger task decompositions and more complex specifications, as summarized in Table~\ref{tab:scenes}.
As expected, O3 and O4 generally yield lower success rates than O1 and O2.
Nevertheless, \textit{Full} continues to achieve high verified success rates on these more complex objectives.
Together with the small gap between generation and verified success, these results show that the framework remains effective as mission complexity increases while largely preserving the original mission requirements.

\noindent\textbf{Answer to RQ1.}
Our framework achieves high success rates across heterogeneous CPS scenes, including on more complex objectives.
Moreover, the small gap between generation and verified success shows that successfully generated controls generally satisfy the original natural language mission requirements.

\begin{table*}[!t]
\centering
\footnotesize
\caption{Ablation of hierarchical feedback: success rates and average runtime.}
\label{tab:strategy-matrix}
{
\setlength{\tabcolsep}{1.5pt}
\renewcommand{\arraystretch}{0.95}
\begin{tabular}{@{}ll*{8}{r}@{}}
\toprule
\textbf{Scene} & \textbf{Model} &
\multicolumn{2}{c}{\textbf{Full}} &
\multicolumn{2}{c}{\textbf{No-TSTL-Check}} &
\multicolumn{2}{c}{\textbf{No-Replan}} &
\multicolumn{2}{c}{\textbf{Simple-Feedback}} \\
\cmidrule(lr){3-4}\cmidrule(lr){5-6}\cmidrule(lr){7-8}\cmidrule(l){9-10}
& & \shortstack{\textbf{Success}\\\textbf{rate}} & \textbf{Time(s)} &
\shortstack{\textbf{Success}\\\textbf{rate}} & \textbf{Time(s)} &
\shortstack{\textbf{Success}\\\textbf{rate}} & \textbf{Time(s)} &
\shortstack{\textbf{Success}\\\textbf{rate}} & \textbf{Time(s)} \\
\midrule
\multirow{3}{*}{scene1} & DeepSeek-V4-Flash & 55\% & 426.1 & \textbf{60\%} & 426.3 & 15\% & 250.6 & 30\% & 358.6 \\
 & GPT-5 & \textbf{85\%} & 528.4 & \textbf{85\%} & 446.9 & 45\% & 249.9 & 50\% & 323.5 \\
 & Qwen3.7-Plus & \textbf{85\%} & 787.1 & \textbf{85\%} & 612.6 & 60\% & 499.2 & 75\% & 649.4 \\
\midrule
\multirow{3}{*}{scene2} & DeepSeek-V4-Flash & \textbf{80\%} & 406.7 & 45\% & 308.4 & 40\% & 261.9 & 70\% & 424.1 \\
 & GPT-5 & \textbf{80\%} & 474.3 & 50\% & 353.1 & 15\% & 311.5 & 55\% & 482.3 \\
 & Qwen3.7-Plus & \textbf{75\%} & 843.1 & 65\% & 905.0 & 15\% & 535.4 & 35\% & 1008.7 \\
\midrule
\multirow{3}{*}{scene3} & DeepSeek-V4-Flash & \textbf{80\%} & 178.8 & 75\% & 194.5 & 55\% & 157.3 & 50\% & 203.6 \\
 & GPT-5 & \textbf{80\%} & 193.7 & 65\% & 189.4 & 60\% & 149.8 & 75\% & 208.6 \\
 & Qwen3.7-Plus & \textbf{25\%} & 438.0 & \textbf{25\%} & 339.8 & 20\% & 287.2 & \textbf{25\%} & 270.7 \\
\midrule
\multirow{3}{*}{scene4} & DeepSeek-V4-Flash & 75\% & 181.1 & \textbf{85\%} & 216.1 & 75\% & 177.8 & \textbf{85\%} & 199.9 \\
 & GPT-5 & \textbf{90\%} & 307.5 & 60\% & 256.9 & 50\% & 170.0 & 50\% & 247.2 \\
 & Qwen3.7-Plus & \textbf{95\%} & 393.3 & 90\% & 412.0 & 85\% & 300.7 & 75\% & 406.0 \\
\midrule
\multirow{3}{*}{scene5} & DeepSeek-V4-Flash & 50\% & 185.3 & \textbf{60\%} & 199.4 & 15\% & 156.5 & 25\% & 191.7 \\
 & GPT-5 & \textbf{85\%} & 334.0 & 75\% & 287.9 & 30\% & 181.1 & 30\% & 243.4 \\
 & Qwen3.7-Plus & \textbf{95\%} & 535.4 & \textbf{95\%} & 536.5 & 0\% & -- & 0\% & -- \\
\midrule
\multirow{3}{*}{scene6} & DeepSeek-V4-Flash & \textbf{100\%} & 296.7 & 95\% & 218.1 & 40\% & 185.3 & 95\% & 286.4 \\
 & GPT-5 & \textbf{95\%} & 196.9 & \textbf{95\%} & 197.3 & \textbf{95\%} & 181.0 & \textbf{95\%} & 186.2 \\
 & Qwen3.7-Plus & \textbf{100\%} & 451.4 & 90\% & 508.2 & 75\% & 428.3 & \textbf{100\%} & 457.7 \\
\midrule
\multirow{3}{*}{scene7} & DeepSeek-V4-Flash & 95\% & 252.0 & 95\% & 251.9 & 55\% & 208.1 & \textbf{100\%} & 322.1 \\
 & GPT-5 & \textbf{100\%} & 201.5 & 85\% & 192.5 & 85\% & 162.3 & 90\% & 235.4 \\
 & Qwen3.7-Plus & 95\% & 452.4 & \textbf{100\%} & 423.9 & 95\% & 417.5 & \textbf{100\%} & 481.3 \\
\midrule
\multirow{3}{*}{\textbf{All}} & DeepSeek-V4-Flash & \textbf{76.4\%} & 274.3 & 73.6\% & 250.6 & 42.1\% & 194.7 & 65.0\% & 289.7 \\
 & GPT-5 & \textbf{87.9\%} & 314.3 & 73.6\% & 272.0 & 54.3\% & 183.8 & 63.6\% & 262.7 \\
 & Qwen3.7-Plus & \textbf{81.4\%} & 556.9 & 78.6\% & 537.4 & 50.0\% & 403.0 & 58.6\% & 524.7 \\
\bottomrule
\end{tabular}}
\vspace{-15pt}
\end{table*}

\subsection{Ablation Study of Hierarchical Feedback} 
\label{sec:ablation}
To answer RQ2, we examine how the individual feedback mechanisms contribute to the success of our framework.
Table~\ref{tab:strategy-matrix} compares the four strategies on DeepSeek-V4-Flash, GPT-5, and Qwen3.7-Plus.
Each row reports the average results across four task objectives.
The \textit{success rate} denotes the average task completion rate, and \textit{Time(s)} reports the mean runtime over successful runs.

Regarding overall performance, our framework (strategy \textit{full}) achieves the best aggregate success rate for all three LLM models: 76.4\% on DeepSeek-V4-Flash, 87.9\% on GPT-5, and 81.4\% on Qwen3.7-Plus, outperforming all ablated variants.

Among all strategies, \textit{No-Replan} consumes the least time, which is expected because it performs at most one outer-level planning attempt.
However, it exhibits the lowest success rate on all three models.
Its performance gap relative to \textit{Full} shows the value of reconsidering planning decisions in response to synthesis failures, beyond correcting errors within the initial plan.
\textit{Simple-Feedback} achieves the second lowest success rate, despite retaining the same feedback routing as \textit{Full}.
These results show that simply notifying the LLMs that an error has occurred is not enough.
Localizing the fault and telling the model exactly what went wrong is also critically important.
\textit{No-TSTL-Check} yields comparatively better results.
Nevertheless, it lowers aggregate success on all three models.
The largest improvement from these checks occurs with GPT-5: \textit{Full} gains 14.3 percentage points over the setting without these checks.
These results support early task-level validation as a complement to feedback from control synthesis.

\textbf{Answer to RQ2.}
The results support early task validation and replanning guided by the control synthesis tool as complementary parts of the framework.
Replanning provides opportunities to recover from synthesis failures, while informative feedback makes those opportunities more effective.

\subsection{Comparison with Baselines on AI2-THOR}
\label{sec:ai2thor-comparison}
\begin{table*}[!t]
\centering
\caption{Comparison with LLM-based planning methods on AI2-THOR.}
\footnotesize
\label{tab:ai2thor_comparison}
{
\setlength{\tabcolsep}{1.5pt}
\renewcommand{\arraystretch}{0.92}
\begin{tabular}{@{}lllrrrrrrr@{}}
\toprule
\textbf{Scene} & \textbf{Model} & \textbf{Method} &
\shortstack{\textbf{Time}\\\textbf{(s)}} &
\shortstack{\textbf{Generation}\\\textbf{Success} \textbf{(\%)}} &
\shortstack{\textbf{AI2-THOR}\\\textbf{Success} \textbf{(\%)}} &
\multicolumn{4}{c}{Travel Cost} \\
\cmidrule(l){7-10}
& & & & & & \textbf{O1} & \textbf{O2} & \textbf{O3} & \textbf{O4} \\
\midrule
\multirow{9}{*}{Scene 6}
& \multirow{3}{*}{DeepSeek-V4-Flash} & Our Framework & 320.0 & \textbf{100} & \textbf{95} & 1.77 & \textbf{1.70} & \textbf{2.66} & \textbf{4.49} \\
& & SMART-LLM & 26.8 & \textbf{100} & 10 & \textbf{1.50} & -- & 7.75 & -- \\
& & NL2HLTL2PLAN & 174.5 & 50 & 0 & -- & -- & -- & -- \\
\cmidrule(lr){2-10}
& \multirow{3}{*}{GPT-5} & Our Framework & 243.1 & 95 & \textbf{90} & 1.65 & \textbf{1.70} & \textbf{3.27} & \textbf{4.37} \\
& & SMART-LLM & 110.4 & \textbf{100} & 80 & \textbf{1.50} & 4.50 & 5.00 & 12.88 \\
& & NL2HLTL2PLAN & 259.3 & 85 & 0 & -- & -- & -- & -- \\
\cmidrule(lr){2-10}
& \multirow{3}{*}{Qwen3.7-Plus} & Our Framework & 474.1 & \textbf{100} & \textbf{100} & \textbf{1.84} & \textbf{1.70} & \textbf{2.66} & \textbf{4.31} \\
& & SMART-LLM & -- & 10 & 0 & -- & -- & -- & -- \\
& & NL2HLTL2PLAN & 252.7 & 55 & 0 & -- & -- & -- & -- \\
\midrule
\multirow{9}{*}{Scene 7}
& \multirow{3}{*}{DeepSeek-V4-Flash} & Our Framework & 286.3 & 95 & \textbf{90} & \textbf{1.28} & \textbf{0.10} & \textbf{1.38} & \textbf{1.48} \\
& & SMART-LLM & 30.1 & \textbf{100} & 5 & 3.75 & -- & -- & -- \\
& & NL2HLTL2PLAN & 76.3 & 70 & 0 & -- & -- & -- & -- \\
\cmidrule(lr){2-10}
& \multirow{3}{*}{GPT-5} & Our Framework & 224.1 & \textbf{100} & \textbf{100} & \textbf{1.37} & \textbf{0.10} & \textbf{1.38} & \textbf{2.41} \\
& & SMART-LLM & 106.8 & \textbf{100} & 80 & 3.75 & 4.15 & 5.75 & 5.83 \\
& & NL2HLTL2PLAN & 283.4 & 95 & 0 & -- & -- & -- & -- \\
\cmidrule(lr){2-10}
& \multirow{3}{*}{Qwen3.7-Plus} & Our Framework & 519.5 & \textbf{95} & \textbf{90} & \textbf{1.28} & \textbf{0.10} & \textbf{1.29} & \textbf{1.36} \\
& & SMART-LLM & -- & 10 & 0 & -- & -- & -- & -- \\
& & NL2HLTL2PLAN & 265.0 & 30 & 0 & -- & -- & -- & -- \\
\bottomrule
\end{tabular}
}
\vspace{-10pt}
\end{table*}

This subsection answers RQ3 by comparing our framework with two representative LLM-based task planning works, SMART-LLM~\cite{kannan_smartllm_2024} and NL2HLTL2PLAN~\cite{xu_nl2hltl2plan_2025}.
We report four evaluation metrics.
\emph{Time (s)} measures the time from method invocation to completion of the final output.
\emph{Generation success} follows each method's own criterion: successful output reported by the underlying tool for our framework and NL2HLTL2PLAN, and production of Python code for SMART-LLM.
\emph{AI2-THOR success} indicates whether the generated outputs can be executed successfully in the AI2-THOR simulator.
\emph{Travel cost} is the total movement distance across all agents, averaged over successful executions for each objective.

\begin{figure}[!t]
  \centering
  \vspace{-0.3cm}
  \includegraphics[width=0.52\textwidth]{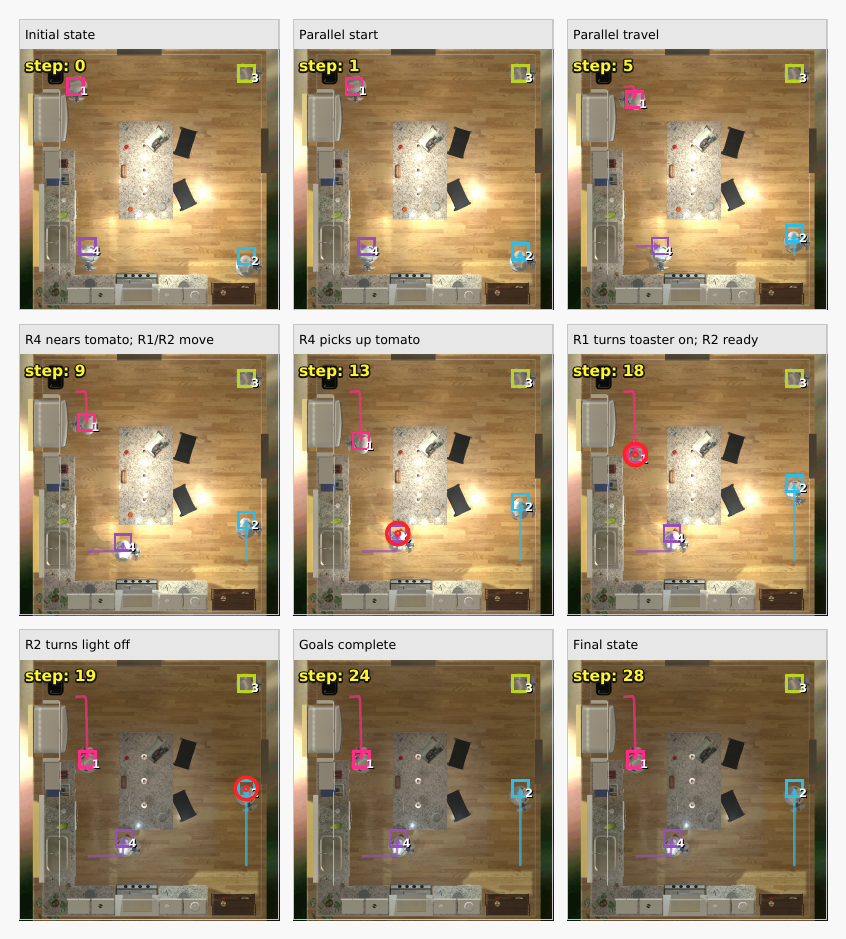}
  \vspace{-0.55cm}
  \caption{Successful execution of a Scene~6 mission in AI2-THOR.}
  \label{ai2thor}
  \Description{Successful Scene 6 execution generated by our method. }
  \vspace{-0.6cm}
\end{figure}

Table~\ref{tab:ai2thor_comparison} shows that our framework not only achieves a high generation success rate but also attains a higher execution success in real simulations within AI2-THOR.
Moreover, the travel cost further demonstrates the efficiency of our planning approach.
Although our method consumes a relatively large amount of time, mainly due to the multiple rounds of LLM calls, this additional cost is justified by the high success rate achieved during actual execution.
Figure~\ref{ai2thor} illustrates a successful execution in Scene~6.

SMART-LLM exhibits substantial variation in success rate across the evaluated models.
With DeepSeek-V4-Flash and Qwen3.7-Plus, some generated responses contain explanatory text or Markdown formatting that prevents direct Python execution in its evaluated pipeline.
Most runs with Qwen3.7-Plus also exceed the default 600-second request timeout.
With GPT-5, SMART-LLM achieves its highest execution success rates among the evaluated models, but remains below our framework in both scenes.
Its travel costs, averaged over successful executions, are also higher for most objectives.

NL2HLTL2PLAN achieves generation success under its internal grid abstraction, but none of the resulting plans lead to a successful AI2-THOR execution in our experiments.
In our integration, we observe failures when translating the generated grid trajectories into executable simulator actions.
This discrepancy highlights the distinction between success within the planning abstraction and successful execution in the target environment.

\textbf{Answer to RQ3.}
On the two evaluated household scenes, our framework achieves higher end-to-end simulator execution success than the tested baselines.
The comparison supports its applicability to these multi-robot tasks, while maintaining accuracy and reliability.

\subsection{Threats to Validity}
LLM hallucination is an inherent threat to validity, as it may introduce syntactic errors in generated outputs or inconsistencies between task plans and STLs.
Nevertheless, we observe that the majority of such errors are successfully repaired under our iterative feedback loop.
The main source of failure is that the final plan still violates some complex control-level temporal constraints after all repair attempts are exhausted.

We also identify another category of failures.
Consider the motivating example, which requires two trucks to move in parallel toward their respective targets.
The LLMs initially decompose the objective into two parallel subtasks, but upon receiving collision constraint violation reports from the control synthesis tool, they may restructure the plan by serializing the two subtasks — unintentionally abandoning the "in parallel" requirement of the original natural language objective.
In fact, the resulting plan and trajectory are physically safe and executable, and the control synthesis is successful and can not afford failure information.
We count such cases as failures under our strict evaluation criterion.
We will further explore how to guide LLMs to generate plans that conform to constraints while better preserving high-level semantic intent in future work.
\section{Related Work}

\noindent\textbf{Temporal logic-based CPS control synthesis and planning.}
Temporal logic has long been used to specify temporal constraints and requirements.
A line of work studies temporal logic falsification, which aims to find counterexamples in systems that violate the given temporal logic specifications.
S-TaLiRo~\cite{liu_staliro_2011} provided an advanced optimization-based CPS falsification tool.
SNIFF~\cite{wang_sniff_2024} designed a scenario-based hierarchical modeling language to formalize reconfigurable CPSs and developed a path-oriented approach to falsify temporal logic specifications for CPSs.

A related class of temporal logic for CPS falsification is temporal logic for planning.
Luo et al.~\cite{luo_plan_2025} developed a search-based approach to synthesize plans for multi-robot systems under temporal logic specifications.
Real-time reactive task allocation and planning under temporal logic specifications have also been studied for large heterogeneous multi-robot systems~\cite{chen_real_2025}.
The above provide strong modeling and planning capabilities, but they typically assume that temporal logic specifications are manually given.
Our work is complementary: it focuses on converting natural language objectives into executable scenario tasks and STL specifications, enabling feasible control synthesis.

\noindent\textbf{LLMs for generating plans and temporal logic specifications.}
Much existing work has used large language models (LLMs) to convert natural language instructions into plans or high-level action sequences~\cite{huang_planners_2022, ichter_saycan_2023, kannan_smartllm_2024, lin_text2motion_2023}.
For example, SMART-LLM~\cite{kannan_smartllm_2024} explored the use of LLMs in performing task decomposition and allocation for multi-robot tasks.
Its final output is executable code.
Other approaches tend to leverage LLMs to generate structured representations at the first stage.
NL2HLTL2PLAN~\cite{xu_nl2hltl2plan_2025} used LLMs to translate natural language instructions into a hierarchical task tree to capture logical relations.
SELP~\cite{wu_selp_2025} developed a method that uses LLMs to generate LTL specifications from natural language, selects the candidate, and then uses LLMs to produce a plan that conforms to it.
AutoTAMP~\cite{chen_autotamp_2024} generates STLs that describe objectives, using LLMs as translators and checkers.

Beyond planning, DeepSTL~\cite{he_deepstl_2022}, NL2TL~\cite{chen_nl2tl_2023}, KGST~\cite{fang_kgst_2025}, LTLCodeGen~\cite{rabiei_ltlcodegen_2025}, and so on studied natural language to temporal logic translation.
These works focus on generating temporal logic.
Our framework differs by jointly generating a hierarchical scenario task, task-level and control-level STL from LLMs, and finally synthesizing feasible control for multi-component CPSs.

\noindent\textbf{LLMs for multi-agent planning and coordination.}
Recent work also explores LLMs as coordinators for multi-agent systems.
RoCo~\cite{zhao_roco_2024} proposed a multi-robot collaboration method.
Robots are equipped with LLMs to discuss and collectively reason about strategies and also get feedback from the environment.
LLM-MARS~\cite{lykov_llmmars_2023} enables dialogues between humans and robots, allowing them to generate behavior based on operator commands.
Chen et al.~\cite{chen_collaboration_2024} compare centralized, decentralized, and hybrid LLM-based communication architectures for multi-robot collaboration in terms of task success and token efficiency.
These studies show the potential of LLMs for communication and coordination among agents and robots, but they are not primarily designed to produce constraints and specifications that agents should conform to.

\noindent\textbf{Feedback guided LLM planning and repair.}
Prompting enables LLMs to perform new tasks from natural language instructions, examples, and contextual information~\cite{brown_gpt3_2020}.
Besides, re-prompting LLMs with additional context from the environment, execution, or validation feedback has been shown to improve task planning~\cite{huang_monologue_2023, guan_leveraging_2023}.
Raman et al.~\cite{raman_reprompt_2022} used re-prompting with precondition errors from failed actions to obtain more executable plans.
Silver et al.~\cite{silver_pddl_2024} re-prompt LLMs with feedback to repair Python programs synthesized as generalized planners.
Skreta et al.~\cite{skreta_errors_2023} use iterative prompting to incorporate verifier-reported errors into subsequent prompts, improving the validity of generated task programs.
AutoTAMP~\cite{chen_autotamp_2024} uses syntax checking and planner-generated state sequences to correct STL translations.
Our framework uses control synthesis diagnostics to guide revisions of the task plan itself.
We extend the underlying tool to identify violated constraints, conflicting component assignments, and failures involving component guards or invariants, together with the associated tasks.
This feedback allows the LLM to reconsider task decomposition and component allocation in response to failures arising from component dynamics and interactions.

\section{Conclusion}
This paper presents a hierarchical feedback framework for generating executable plans for multi-component CPSs from natural language missions.
A consistency checking algorithm detects contradictions between LLM-generated task plans and STL specifications before control synthesis.
When synthesis fails, diagnostic information guides revisions to task decomposition and allocation.
Experimental results demonstrate that early consistency checking and execution-guided replanning complement each other in improving execution success, and that informative diagnostics make replanning more effective.

\bibliographystyle{ACM-Reference-Format}
\bibliography{mybibliography}

\end{document}